\documentclass{article}

\usepackage[final]{neurips_2023}

\usepackage[utf8]{inputenc}
\usepackage[T1]{fontenc}
\usepackage{hyperref}
\usepackage{url}
\usepackage{booktabs}
\usepackage{amsfonts}
\usepackage{amsmath}
\usepackage{nicefrac}
\usepackage{microtype}
\usepackage{xcolor}
\usepackage{graphicx}
\usepackage{algorithm}
\usepackage{algorithmic}
\usepackage{enumitem}
\usepackage{tikz}
\usetikzlibrary{arrows.meta, positioning, shapes.geometric, calc, decorations.pathreplacing}
\usepackage{listings}
\usepackage{multirow}
\usepackage{pifont}
\usepackage{pgfplots}
\pgfplotsset{compat=1.18}
\usepgfplotslibrary{units}
\newcommand{\cmark}{\ding{51}}
\newcommand{\xmark}{\ding{55}}

\title{RightNow-Arabic-0.5B-Turbo: An Open Sub-1B Arabic Language Model via Vocabulary Injection and Edge-First Deployment}

\author{
  Jaber Jaber\thanks{Correspondence: \texttt{jaber@rightnowai.co}} \\
  RightNow AI\\
  \texttt{jaber@rightnowai.co} \\
  \And
  Osama Jaber \\
  RightNow AI\\
  \texttt{osama@rightnowai.co} \\
}

\begin{document}
\maketitle

\begin{center}
\includegraphics[height=1.1cm]{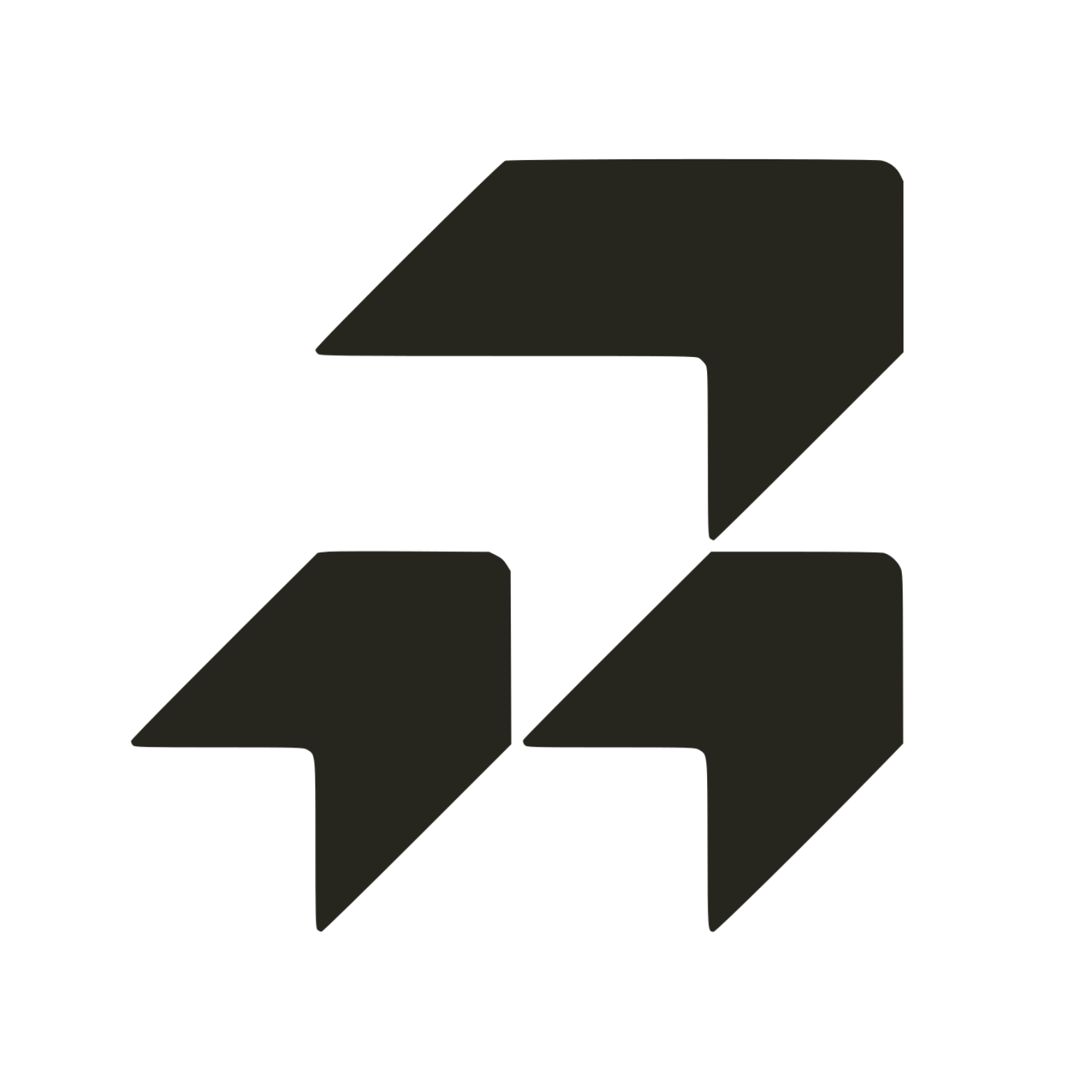}
\end{center}
\vspace{-0.3em}

\begin{abstract}
Open Arabic large language models split into two classes: sub-1B multilingual models that treat Arabic as an afterthought (Qwen2.5-0.5B, Falcon-H1-0.5B), and 7B--70B Arabic-specialized models that require a server to run (Jais, AceGPT, ALLaM, SILMA). The one published attempt at a sub-2B Arabic-specialized model, Kuwain-1.5B, never released its weights. We present \textsc{RightNow-Arabic-0.5B-Turbo}, a 518M-parameter Arabic-specialized decoder LLM built on Qwen2.5-0.5B. The pipeline adds 27{,}032 Arabic tokens via mean-subtoken initialization, continues pretraining on 504M Arabic tokens on $8\times$H100 with FSDP, FlashAttention varlen packing, and Liger fused kernels, then applies supervised fine-tuning on 129{,}116 Arabic instruction pairs with response-only loss masking, direct preference optimization on 6{,}750 Arabic preference pairs, and weight soup merging across three checkpoints. On three lm-evaluation-harness Arabic benchmarks (COPA-ar, Arabic HellaSwag, ArabicMMLU) the merged model reaches 35.9\% mean accuracy, beats every same-class open model, ties Falcon-H1-1.5B on COPA-ar (58.4\%) at one-third the size, and recovers 67\% of SILMA-9B's mean at 1/18 the parameters. The edge build quantizes to 398\,MB (q4\_k\_m) and delivers 635 tokens/s at batch size 1 on a single H100 via llama.cpp. All code (5{,}555 lines across 25 scripts), weights (bf16, int8, and four GGUF quantizations), and benchmark scripts are released openly at \url{https://huggingface.co/RightNowAI/RightNow-Arabic-0.5B-Turbo}.
\end{abstract}

\section{Introduction}
\label{sec:intro}

Arabic is spoken by more than 400 million people and is the official language of 25 countries, yet open Arabic language model weights trail English models by an uncomfortable margin. The strongest open Arabic models, Jais \citep{sengupta2023jais}, AceGPT \citep{huang2023acegpt}, ALLaM \citep{bari2024allam}, and SILMA, all live in the 7B--70B parameter range. They achieve strong benchmark scores but demand 16--140\,GB of memory and are out of reach for phone, laptop-CPU, and embedded deployment. At the other end of the spectrum, the sub-1B open models that can actually run on commodity hardware, Qwen2.5-0.5B-Instruct \citep{qwen25}, Falcon-H1-0.5B, treat Arabic as one of a hundred languages and do not allocate Arabic-specialized vocabulary or training data.

The one published effort that targeted this gap, Kuwain-1.5B \citep{hennara2024kuwain}, proposed ``language injection'' on TinyLlama \citep{zhang2024tinyllama} and reported an 8\% average Arabic improvement over the base. Misraj.ai (Khobar, Saudi Arabia) released the paper but did not release the weights. This leaves practical Arabic deployment on edge devices without a real option: the smallest downloadable Arabic-specialized decoder LLM on HuggingFace is nothing.

We present \textsc{RightNow-Arabic-0.5B-Turbo}, a 518M-parameter Arabic-specialized decoder LLM. The model is built by injecting 27{,}032 new Arabic tokens into Qwen2.5-0.5B \citep{qwen25}, continuing pretraining on 504M Arabic tokens, supervised fine-tuning with response-only loss masking on 129{,}116 instruction pairs, applying direct preference optimization \citep{rafailov2023dpo}, and merging three checkpoints via weight averaging \citep{wortsman2022soups}. The result is the smallest open Arabic-specialized decoder LLM released to date. On COPA-ar it matches Falcon-H1-1.5B at one-third the parameter count; quantized to 4 bits it fits in 398\,MB and reaches 635 tokens/s at bs=1 on a single H100 via llama.cpp.

\paragraph{Key insight.}
Sub-1B Arabic models do not need new training techniques, every component of our pipeline exists in prior work. What they need is careful orchestration: vocabulary expansion that actually lowers Arabic token fertility, a data loader that avoids multi-rank HuggingFace Hub stalls, SFT that masks prompt tokens so the loss signal is concentrated on the assistant response, and an export path that produces artifacts small enough for the device class the model is targeting. Every mistake costs the accuracy budget of a model already near its scale ceiling.

\paragraph{Contributions.}
\begin{enumerate}[leftmargin=*, itemsep=0.15em, topsep=0.2em]
\item The smallest open Arabic-specialized decoder LLM on HuggingFace as of the submission date, at 518M parameters and 398\,MB on disk (q4\_k\_m).
\item A reproducible vocabulary-injection pipeline that mean-subtoken-initializes 27{,}032 Arabic tokens into Qwen2.5-0.5B and cuts Arabic tokenizer fertility from 2.18 to 1.80 tokens per word (17.3\% reduction).
\item Direct head-to-head benchmarks against 6 competing models on three Arabic lm-evaluation-harness tasks under identical methodology, showing where a 0.5B model wins and where it loses to models 3--18$\times$ larger.
\item A weight-soup merge ablation over 7 variants (SLERP and LERP across the DPO/SFT/pretrain checkpoints), selecting a configuration that improves mean accuracy by 0.44 absolute points over the DPO endpoint alone.
\item llama.cpp GGUF conversion producing 4 quantization levels (f16, q8\_0, q5\_k\_m, q4\_k\_m) with measured throughput of 582--646 tokens/s at bs=1 on H100.
\item Full open release: 5{,}555 lines of Python across 25 scripts and 13 runtime modules, all training configs, all benchmark scripts, all intermediate checkpoints, and the final weights.
\end{enumerate}

\section{Related Work}
\label{sec:related}

\paragraph{Arabic-specialized LLMs.}
Jais \citep{sengupta2023jais} from Inception/MBZUAI set the initial standard with a GPT-3-style 13B/30B decoder pretrained on a mixture of Arabic, English, and code. AceGPT \citep{huang2023acegpt} from FreedomIntelligence targeted cultural alignment via RLAIF and released 7B/13B chat variants. ALLaM \citep{bari2024allam} from SDAIA (Saudi Arabia) scaled to 7B/13B/34B/70B using vocabulary expansion and Arabic-English mixed pretraining. SILMA is a 9B Arabic-specialized model released by SILMA AI without an accompanying paper. All four operate in the 7B+ parameter range and target server deployment.

\paragraph{Sub-2B Arabic models.}
The closest prior work to ours is Kuwain-1.5B \citep{hennara2024kuwain}, published by Misraj.ai in April 2025. Kuwain injects Arabic vocabulary into TinyLlama-1.1B \citep{zhang2024tinyllama} and reports an average 8\% improvement on Arabic benchmarks over the base. Mutarjim \citep{misraj2025mutarjim} builds on Kuwain for Arabic-English translation. Critically, neither Kuwain nor Mutarjim weights are published on HuggingFace at the time of this submission, so neither can be directly compared or deployed. RightNow-Arabic-0.5B-Turbo is strictly smaller (518M vs 1.5B) and fully open.

\paragraph{Vocabulary expansion for language adaptation.}
Extending a pretrained tokenizer with new vocabulary is a standard adaptation technique. ALLaM \citep{bari2024allam} argues that vocabulary expansion paired with English anchoring prevents catastrophic forgetting. The specific initialization scheme we use, averaging the old embeddings of a new token's sub-piece decomposition, was formalized by WECHSEL \citep{minixhofer2022wechsel} and subsequently adopted by Kuwain and ALLaM.

\paragraph{Small-model training stack.}
Training on 8$\times$H100 SXM5 demands specific infrastructure. FlashAttention \citep{dao2022flashattention} and its varlen variant remove the memory and kernel-launch overhead of dense attention. PyTorch FSDP \citep{zhao2023fsdp} shards optimizer and gradient state; we use the \texttt{\_HYBRID\_SHARD\_ZERO2} strategy. Liger Kernel \citep{hsu2024liger} replaces RMSNorm, RoPE, SwiGLU, and fused linear cross-entropy with Triton implementations, saving memory and time. Our vocabulary expansion increases the output projection to 178{,}697 rows, which makes fused linear cross-entropy essential: materialized float32 logits at batch 16$\times$4096$\times$178697 would require 44\,GiB.

\paragraph{Post-training.}
Direct preference optimization \citep{rafailov2023dpo} reformulates RLHF as a classification loss over preference pairs, eliminating the reward model. Model souping \citep{wortsman2022soups} averages weights across multiple fine-tuned checkpoints and has been shown to improve out-of-distribution generalization. Our pipeline applies both: DPO on 6{,}750 Arabic preference pairs from argilla-dpo-mix-7k-arabic, and a linear-weight soup across the DPO, SFT, and pretrain checkpoints.

\paragraph{Data.}
Pretraining uses Arabic Wikipedia (504M tokens after merging with the new tokenizer) via the wikimedia/wikipedia corpus. We originally planned to mix in FineWeb-2-ar \citep{penedo2024fineweb} but encountered persistent HuggingFace Hub 504 stalls during multi-rank streaming; we resolved this by pre-tokenizing the corpus to a flat \texttt{int32} memmap and sampling windows per rank. Instruction tuning uses a merge of five Arabic instruction datasets; preference tuning uses the argilla Arabic DPO mix. All data is publicly available. Evaluation uses three tasks from lm-evaluation-harness \citep{gao2021lmeval}: COPA-ar, Arabic MT HellaSwag, and the Arabic Leaderboard ArabicMMLU \citep{koto2024arabicmmlu}. Table~\ref{tab:landscape} summarizes the open Arabic LLM landscape and shows where our model sits.

\begin{table}[t]
\centering
\caption{The open Arabic LLM landscape. Our model is the only sub-1B Arabic-specialized decoder with publicly available weights.}
\label{tab:landscape}
\footnotesize
\setlength{\tabcolsep}{4pt}
\begin{tabular}{@{}lccc@{}}
\toprule
Model & Params & Open weights & Arabic-specialized \\
\midrule
Qwen2.5-0.5B-Instruct \citep{qwen25} & 494M & \cmark & \xmark \\
Falcon-H1-0.5B-Instruct & 524M & \cmark & \xmark \\
\textbf{RightNow-Arabic-0.5B-Turbo (ours)} & \textbf{518M} & \cmark & \cmark \\
Falcon-H1-1.5B-Instruct & 1.5B & \cmark & \xmark \\
Kuwain-1.5B \citep{hennara2024kuwain} & 1.5B & \xmark & \cmark \\
AceGPT-7B-chat \citep{huang2023acegpt} & 7B & \cmark & \cmark \\
ALLaM-7B-Instruct \citep{bari2024allam} & 7B & \cmark & \cmark \\
SILMA-9B-Instruct & 9B & \cmark & \cmark \\
Jais-13B-chat \citep{sengupta2023jais} & 13B & \cmark & \cmark \\
\bottomrule
\end{tabular}
\end{table}

\section{Method}
\label{sec:method}

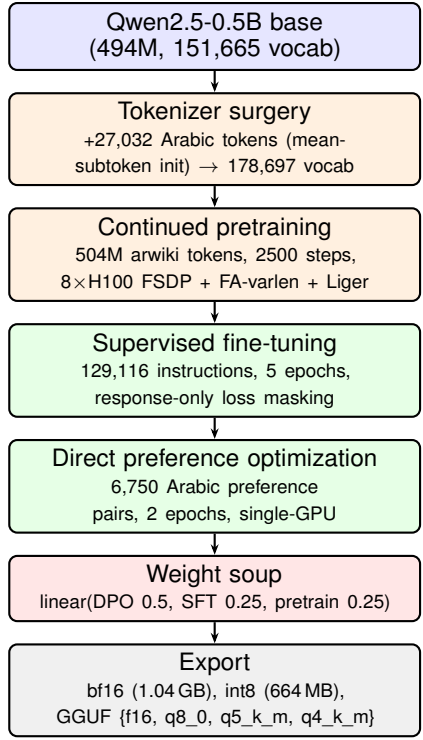
\begin{figure}[t]
\centering
\begin{tikzpicture}[
  box/.style={draw, rounded corners=3pt, thick, minimum height=0.75cm,
              align=center, font=\footnotesize\sffamily, text width=5.2cm},
  arr/.style={-{Stealth[length=4pt]}, thick},
  node distance=0.28cm and 0.8cm,
]
\node[box, fill=blue!10] (base) {Qwen2.5-0.5B base (494M, 151{,}665 vocab)};
\node[box, fill=orange!12, below=of base] (tok) {Tokenizer surgery\\ \scriptsize +27{,}032 Arabic tokens (mean-subtoken init) $\to$ 178{,}697 vocab};
\node[box, fill=orange!12, below=of tok] (pre) {Continued pretraining\\ \scriptsize 504M arwiki tokens, 2500 steps, 8$\times$H100 FSDP + FA-varlen + Liger};
\node[box, fill=green!10, below=of pre] (sft) {Supervised fine-tuning\\ \scriptsize 129{,}116 instructions, 5 epochs, response-only loss masking};
\node[box, fill=green!10, below=of sft] (dpo) {Direct preference optimization\\ \scriptsize 6{,}750 Arabic preference pairs, 2 epochs, single-GPU};
\node[box, fill=red!10, below=of dpo] (soup) {Weight soup\\ \scriptsize linear(DPO 0.5, SFT 0.25, pretrain 0.25)};
\node[box, fill=gray!12, below=of soup] (exp) {Export\\ \scriptsize bf16 (1.04\,GB), int8 (664\,MB), GGUF \{f16, q8\_0, q5\_k\_m, q4\_k\_m\}};
\draw[arr] (base) -- (tok);
\draw[arr] (tok) -- (pre);
\draw[arr] (pre) -- (sft);
\draw[arr] (sft) -- (dpo);
\draw[arr] (dpo) -- (soup);
\draw[arr] (soup) -- (exp);
\end{tikzpicture}
\caption{Training and deployment pipeline. Each stage produces a checkpoint that feeds the next; the soup stage also consumes the pretrain and SFT checkpoints directly.}
\label{fig:pipeline}
\end{figure}

Figure~\ref{fig:pipeline} shows the full pipeline. The model starts as Qwen2.5-0.5B (base, not Instruct), gains 27{,}032 new Arabic tokens, sees 504M pretraining tokens, 129{,}116 SFT examples, and 6{,}750 DPO pairs, and ends as a merged checkpoint exported in bf16, int8, and four GGUF quantizations.

\subsection{Tokenizer surgery}
\label{sec:tokenizer}

Qwen2.5-0.5B's original byte-level BPE tokenizer has 151{,}665 tokens and encodes Arabic text at 2.18 tokens per word on a held-out sample, roughly $1.4\times$ the English rate. To lower this, we train a SentencePiece unigram model \citep{kudo2018sentencepiece} with 32{,}000 tokens on a 12.54\,GB Arabic corpus composed of 21.5M lines of Arabic Wikipedia (via wikimedia/wikipedia 20231101.ar) and 5.4\,GB of filtered Arabic web text. We apply standard Arabic text normalization before training: NFKC, alif-variant normalization (hamza-above, hamza-below, madda $\to$ bare alif), tatweel stripping, and ya-variant normalization.

The resulting SentencePiece model is merged into the Qwen tokenizer with a deduplication pass: any SentencePiece piece whose surface string already round-trips to a single existing Qwen token is discarded, leaving 27{,}032 net-new Arabic tokens. We then call \texttt{model.resize\_token\_embeddings(178697)} and initialize the 27{,}032 new embedding rows as in Algorithm~\ref{alg:meaninit}.

\begin{algorithm}[t]
\caption{Mean-subtoken embedding initialization}
\label{alg:meaninit}
\begin{algorithmic}[1]
\REQUIRE old tokenizer $T_{\text{old}}$, new token strings $\mathcal{N}$, embedding matrix $E \in \mathbb{R}^{V_{\text{new}} \times d}$, original vocabulary size $V_{\text{old}}$
\STATE $E_{\text{old}} \leftarrow E[:V_{\text{old}}].\text{clone}()$ \hfill \textit{// freeze original rows}
\FORALL{new token id $n$ with surface string $s$ in $\mathcal{N}$}
  \STATE $I \leftarrow T_{\text{old}}.\text{encode}(s, \text{add\_special\_tokens}=\text{False})$
  \IF{$|I| = 0$} \STATE $I \leftarrow [T_{\text{old}}.\text{unk\_id}]$ \ENDIF
  \STATE $E[n] \leftarrow \frac{1}{|I|}\sum_{i \in I} E_{\text{old}}[i]$
\ENDFOR
\IF{\textsc{lm\_head.weight.data\_ptr} $=$ \textsc{embed.weight.data\_ptr}}
  \STATE \textit{// tied embeddings: lm\_head rows update automatically}
\ELSE
  \STATE $E_{\text{lm\_head}}[V_{\text{old}}:V_{\text{new}}] \leftarrow E[V_{\text{old}}:V_{\text{new}}]$
\ENDIF
\end{algorithmic}
\end{algorithm}

After merging, we verify that \texttt{model.lm\_head.weight.data\_ptr()} still equals \texttt{model.get\_input\_embeddings().weight.data\_ptr()}, Qwen2.5-0.5B ties these, so that both the input embeddings and output projection receive the mean-init rows simultaneously. On a 1000-word Arabic sample, fertility drops from 2.08 to 1.77 tokens per word (14.7\% reduction). On a larger held-out 368-word sample used for the final measurement, fertility drops from 2.18 to 1.80 (17.3\% reduction).

\subsection{Continued pretraining}
\label{sec:pretrain}

The merged model is trained for 2{,}500 optimizer steps on a pre-tokenized flat \texttt{int32} memmap containing 504{,}318{,}692 tokens of Arabic Wikipedia encoded with the new tokenizer. We use a per-GPU micro-batch of 16 sequences of 4096 tokens with gradient accumulation of 8, giving an effective batch of $16 \times 4096 \times 8 \times 8 = 4.19$M tokens per step and 10.5B total training tokens. FSDP \citep{zhao2023fsdp} wraps each Qwen2DecoderLayer with \texttt{ShardingStrategy.\_HYBRID\_SHARD\_ZERO2} and bf16 mixed precision; FlashAttention varlen \citep{dao2022flashattention} takes packed \texttt{cu\_seqlens} directly from the memmap iterator so document boundaries are respected without padding; Liger fused kernels \citep{hsu2024liger} cover RMSNorm, RoPE, SwiGLU, and fused linear cross-entropy.

The optimizer is fused AdamW, $\beta = (0.9, 0.95)$, $\epsilon = 10^{-8}$, weight decay 0.1, peak learning rate $2\times10^{-4}$ with 500-step linear warmup and cosine decay to $2\times10^{-5}$. Training runs at 415{,}000 tokens/s aggregate on 8$\times$H100 SXM5 with peak per-GPU memory of 24\,GB.

\subsection{Supervised fine-tuning}
\label{sec:sft}

We merge five Arabic instruction datasets into a single deduplicated pool: FreedomIntelligence/evol-instruct-arabic (59{,}022 rows), FreedomIntelligence/alpaca-gpt4-arabic (49{,}969), FreedomIntelligence/sharegpt-arabic (5{,}231), arbml/CIDAR (10{,}000), and the Arabic subset of CohereForAI/aya\_dataset (4{,}947 after filtering on \texttt{language\_code=="arb"}). MD5 deduplication on the rendered ChatML string yields 129{,}116 unique examples.

Each example is rendered into ChatML with the system prompt \textit{``you are a smart assistant that answers in formal Arabic''} and pre-tokenized into two parallel \texttt{int32} memmaps: \texttt{sft\_tokens.bin} contains the full input id sequence, \texttt{sft\_labels.bin} contains $-100$ on prompt positions and the actual token id on assistant-response positions. Training is 5 epochs at peak learning rate $2\times10^{-5}$, grad accumulation 2, 30 warmup steps, and computes loss \emph{only on response tokens} \citep{ouyang2022instructgpt}. Of the 43{,}918{,}266 total pretokenized tokens, 31{,}678{,}374 (72.1\%) carry non-negative-100 labels; masking the prompt saves capacity for the distribution we actually care about.

\subsection{Direct preference optimization}
\label{sec:dpo}

The SFT checkpoint undergoes DPO \citep{rafailov2023dpo} on 6{,}750 Arabic preference pairs from 2A2I/argilla-dpo-mix-7k-arabic. Each pair provides a user prompt, a chosen assistant response, and a rejected assistant response. Training runs on a single H100 with per-device batch size 2, gradient accumulation 8, $\beta = 0.1$, peak learning rate $5\times10^{-7}$, 2 epochs, and an explicit frozen reference model (not LoRA-implicit). Total training time is 34 minutes for 844 optimizer steps.

\subsection{Weight soup merging}
\label{sec:merge}

The DPO run converges to a loss near $\ln 2 \approx 0.693$ with reward-accuracy 0.48 and margin $-6.9\times10^{-4}$, indicating the preference dataset supplies weak signal at this scale. We mitigate this by averaging the DPO checkpoint with the earlier pretrain and SFT checkpoints, all three share the same 178{,}697-vocab architecture, which makes direct weight averaging safe. We produce 7 merge variants: linear interpolations of DPO and pretrain at $t \in \{0.3, 0.5, 0.7\}$, SLERP at $t \in \{0.3, 0.5\}$, linear DPO/SFT at $t=0.5$, and a 50/25/25 soup of DPO/SFT/pretrain. Each merge is benchmarked on the same 3 Arabic tasks (Section~\ref{sec:results}); the 50/25/25 soup wins and becomes the final checkpoint.

\subsection{Edge deployment}
\label{sec:gguf}

For edge deployment we convert the merged model to llama.cpp GGUF format and produce four quantization levels. The llama.cpp converter does not recognize our tokenizer hash (because we added 27k tokens), so we patched \texttt{convert\_hf\_to\_gguf.py} to map our hash to the existing \texttt{qwen2} pre-tokenizer type. After conversion, \texttt{llama-quantize} produces q8\_0 (525\,MB), q5\_k\_m (419\,MB), and q4\_k\_m (398\,MB). Note that q5\_k\_m and q4\_k\_m fall back to higher-bit quantization for 144 of the 290 tensors, specifically the expanded Arabic embedding rows, because k-quants require tile sizes that the added vocabulary does not align with. Effective bits-per-weight is therefore 6.79 (q5\_k\_m) and 6.45 (q4\_k\_m) rather than the nominal 5 and 4.

\section{Implementation}
\label{sec:impl}

Table~\ref{tab:code} summarizes the codebase. Total Python source is 5{,}555 lines across 25 scripts and 13 runtime modules. The full pipeline runs on a single Nebius gpu-h100-sxm instance (8$\times$H100 80GB SXM5, Ubuntu 24.04, CUDA 13.0, PyTorch 2.11.0+cu130, flash-attn 2.8.3, transformers 5.5.0, trl 1.0.0, liger-kernel 0.7.0).

\begin{table}[t]
\centering
\caption{Codebase metrics. Every training, benchmark, merge, and export step is reproducible from the released scripts.}
\label{tab:code}
\footnotesize
\begin{tabular}{@{}lrl@{}}
\toprule
Component & Lines & Notes \\
\midrule
Phase scripts (00--07) & 1{,}904 & env check, tokenizer surgery, data, train, eval \\
Runtime modules \texttt{src/rnar/} & 1{,}468 & FSDP, FA patch, data loader, train loop, merger \\
Benchmark + merge scripts & 1{,}047 & competitor bench, merge variants, speed tests, GGUF \\
Pre-tokenization scripts & 318 & flat memmap producers for pretrain and SFT \\
Chart generation & 524 & matplotlib Pareto and per-task bar plots \\
Configs & 171 & YAML hyperparameters for pretrain, SFT, DPO \\
Utilities + audit & 123 & tokenizer hash patch, paper audit, helpers \\
\midrule
\textbf{Total} & \textbf{5{,}555} & \\
\bottomrule
\end{tabular}
\end{table}

\paragraph{Memmap data loader.}
Early runs attempted HuggingFace Hub streaming for the pretraining corpus. Under 8-rank torchrun, one or more ranks would stall for 15+ minutes at the first parquet fetch, causing NCCL collective deadlocks. We resolved this by pre-tokenizing the entire corpus to a flat int32 memmap (\texttt{arwiki\_tokens.bin}, 2.0\,GB) and sampling fixed-length windows per rank. The loader is 162 lines and serves data at the speed of sequential memory reads, effectively zero overhead compared to the GPU forward pass.

\paragraph{Packed attention with document boundaries.}
The loader also produces \texttt{cu\_seqlens} pointers for every batch: each window is scanned for EOS positions and boundaries are inserted so FlashAttention varlen correctly masks cross-document attention. We monkey-patch \texttt{Qwen2Attention.forward} to read \texttt{cu\_seqlens} from a process-global (not thread-local) dictionary, because PyTorch gradient checkpointing recomputes forwards on a different thread during backward, and a thread-local would be empty at that time.

\paragraph{FSDP hybrid sharding.}
For 0.5B parameters on 8$\times$80\,GB H100, full parameter sharding is unnecessary. We use \texttt{ShardingStrategy.\_HYBRID\_SHARD\_ZERO2}, which replicates parameters across ranks but shards optimizer state and gradients within each rank group. This keeps the training compute local and all-gather overhead minimal.

\section{Experimental Evaluation}
\label{sec:exp}

\paragraph{Hardware.}
All experiments run on a single Nebius \texttt{gpu-h100-sxm} node: 8$\times$NVIDIA H100 80\,GB SXM5 HBM3, NVLink4, 128 vCPUs, 1.5\,TiB RAM, 1.28\,TiB SSD, driver 580.126.09, CUDA 13.0.88. NCCL allreduce busbw was measured at 466.9\,GB/s on a 1\,GiB tensor across 8 ranks at session start.

\paragraph{Evaluation methodology.}
We use lm-evaluation-harness \citep{gao2021lmeval} v0.4.11 with three tasks: \texttt{copa\_ar}, \texttt{arabic\_mt\_hellaswag}, and \texttt{arabic\_leaderboard\_arabic\_mmlu} (our model gets evaluated on the full 14{,}575-question ArabicMMLU \citep{koto2024arabicmmlu}). Every model is scored with \texttt{apply\_chat\_template=True} (with a fallback to raw prompts for models whose tokenizer has no \texttt{chat\_template} attribute, specifically AceGPT-7B-chat), batch size 2, \texttt{max\_length=1536}, per-task \texttt{limit=200}, and \texttt{acc\_norm} preferred over \texttt{acc} where available. The same methodology is applied to all baselines. Evaluating each model on all three tasks takes 10--25 minutes depending on model size.

\subsection{Main results}
\label{sec:results}

\begin{figure}[t]
\centering
\begin{tikzpicture}
\begin{semilogxaxis}[
  width=0.92\linewidth,
  height=6.8cm,
  xlabel={Parameters},
  ylabel={Mean Arabic accuracy (\%)},
  xmin=0.35, xmax=13,
  ymin=28, ymax=57,
  xtick={0.5, 1, 2, 5, 10},
  xticklabels={0.5B, 1B, 2B, 5B, 10B},
  ytick={30, 35, 40, 45, 50, 55},
  grid=major,
  grid style={gray!25, thin},
  tick label style={font=\small},
  label style={font=\small\bfseries},
  legend style={
    at={(0.98,0.02)},
    anchor=south east,
    font=\scriptsize,
    draw=gray!40,
    fill=white,
    row sep=1pt,
  },
  % Edge-deployable zone
  extra x ticks={1.0},
  extra x tick style={grid=major, grid style={blue!15, line width=6pt}},
  extra x tick labels={},
  clip=false,
]

% Shaded edge zone
\fill[blue!4] (axis cs:0.35,28) rectangle (axis cs:1.0,57);
\node[font=\tiny\sffamily, blue!50!black, rotate=90, anchor=south] at (axis cs:0.38,42) {edge-deployable};

% Data points
% 0.5B class (gray circles)
\addplot[only marks, mark=*, mark size=3pt, gray!60, mark options={fill=gray!40}]
  coordinates {(0.494, 34.1) (0.524, 30.7)};

% Ours (red, larger, black edge)
\addplot[only marks, mark=*, mark size=4.5pt, red!70!black, mark options={fill=red!70!black, draw=black, line width=0.6pt}]
  coordinates {(0.518, 35.9)};

% 1.5B (dark gray)
\addplot[only marks, mark=*, mark size=3pt, gray!80!black, mark options={fill=gray!60}]
  coordinates {(1.5, 39.5)};

% 7B Arabic-specialized (navy)
\addplot[only marks, mark=square*, mark size=3pt, blue!60!black, mark options={fill=blue!40!black}]
  coordinates {(7.0, 43.9) (7.0, 49.9)};

% 9B (dark navy)
\addplot[only marks, mark=square*, mark size=3pt, blue!80!black, mark options={fill=blue!70!black}]
  coordinates {(9.0, 53.5)};

% Pareto frontier (dashed)
\addplot[dashed, gray!50, thin] coordinates {
  (0.518, 35.9) (1.5, 39.5) (7.0, 43.9) (7.0, 49.9) (9.0, 53.5)
};

% Labels
\node[font=\scriptsize\bfseries, red!70!black, anchor=south west, xshift=3pt, yshift=2pt]
  at (axis cs:0.518, 35.9) {Ours (0.5B)};

\node[font=\tiny, gray!70!black, anchor=north west, xshift=3pt, yshift=-1pt]
  at (axis cs:0.494, 34.1) {Qwen2.5-0.5B};

\node[font=\tiny, gray!70!black, anchor=north west, xshift=3pt, yshift=-1pt]
  at (axis cs:0.524, 30.7) {Falcon-H1-0.5B};

\node[font=\tiny, gray!70!black, anchor=south west, xshift=4pt, yshift=1pt]
  at (axis cs:1.5, 39.5) {Falcon-H1-1.5B};

\node[font=\tiny, blue!50!black, anchor=north west, xshift=4pt, yshift=-1pt]
  at (axis cs:7.0, 43.9) {AceGPT-7B};

\node[font=\tiny, blue!50!black, anchor=south west, xshift=4pt, yshift=1pt]
  at (axis cs:7.0, 49.9) {ALLaM-7B};

\node[font=\tiny, blue!60!black, anchor=south west, xshift=4pt, yshift=1pt]
  at (axis cs:9.0, 53.5) {SILMA-9B};

% Callout annotation for our model
\draw[-{Stealth[length=3pt]}, red!60!black, thick]
  (axis cs:1.8, 32.5) -- (axis cs:0.56, 35.5);
\node[font=\tiny, red!60!black, text width=3.2cm, align=left, anchor=west]
  at (axis cs:1.9, 32.0) {%
    1/14 size of AceGPT-7B\\
    ties Falcon-H1-1.5B on COPA\\
    398\,MB quantized (q4\_k\_m)};

% Legend
\addlegendimage{only marks, mark=*, red!70!black, mark size=3pt, mark options={draw=black, line width=0.5pt}}
\addlegendentry{RightNow-Arabic (ours)}
\addlegendimage{only marks, mark=*, gray!60, mark size=2.5pt}
\addlegendentry{Multilingual 0.5--1.5B}
\addlegendimage{only marks, mark=square*, blue!50!black, mark size=2.5pt}
\addlegendentry{Arabic-specialized 7--9B}

\end{semilogxaxis}
\end{tikzpicture}
\caption{Mean Arabic benchmark accuracy (COPA-ar, HellaSwag-ar, ArabicMMLU) versus parameter count. Our model is the only Arabic-specialized entry in the sub-1B edge-deployable zone. Dashed line shows the Pareto frontier.}
\label{fig:pareto}
\end{figure}
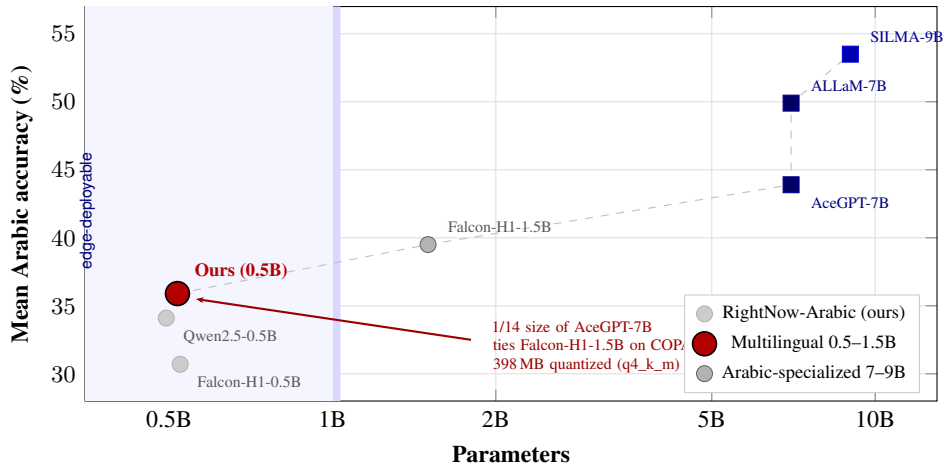

Table~\ref{tab:main} shows the head-to-head comparison. Figure~\ref{fig:pareto} plots the same data on a Pareto scatter of accuracy versus parameters.

\begin{table}[t]
\centering
\caption{Arabic benchmark results. Ours is the only Arabic-specialized open model at 0.5B class. Methodology: lm-eval-harness v0.4.11, \texttt{apply\_chat\_template=True} (fallback for AceGPT-7B-chat which has no template), limit=200 per task, \texttt{acc\_norm} preferred. Bold indicates the best in column; the last column is the unweighted mean.}
\label{tab:main}
\small
\setlength{\tabcolsep}{5pt}
\begin{tabular}{@{}lrrrrr@{}}
\toprule
Model & Params & COPA-ar & HellaSwag-ar & ArabicMMLU & Mean \\
\midrule
\multicolumn{6}{l}{\textit{Same-class open (0.5B)}} \\
Qwen2.5-0.5B-Instruct \citep{qwen25} & 494M & 53.9\% & 22.5\% & \textbf{26.0\%} & 34.1\% \\
Falcon-H1-0.5B-Instruct & 524M & 44.9\% & 23.0\% & 24.2\% & 30.7\% \\
\textbf{Ours (soup\_dpo\_sft\_pre)} & \textbf{518M} & \textbf{58.4\%} & \textbf{26.0\%} & 23.2\% & \textbf{35.9\%} \\
\midrule
\multicolumn{6}{l}{\textit{Larger open}} \\
Falcon-H1-1.5B-Instruct & 1.5B & 58.4\% & 27.5\% & 32.7\% & 39.5\% \\
AceGPT-7B-chat \citep{huang2023acegpt} & 7B & 69.7\% & 27.0\% & 35.0\% & 43.9\% \\
ALLaM-7B-Instruct \citep{bari2024allam} & 7B & 68.5\% & 29.0\% & 52.2\% & 49.9\% \\
SILMA-9B-Instruct & 9B & 69.7\% & 38.0\% & 52.9\% & 53.5\% \\
\bottomrule
\end{tabular}
\end{table}

\paragraph{Same-class wins.}
Among models in the 0.5B class, we are first on COPA-ar (+4.5 vs Qwen2.5-0.5B-Instruct, +13.5 vs Falcon-H1-0.5B), first on HellaSwag-ar (+3.5 vs Qwen, +3.0 vs Falcon), and first on mean (+1.8 vs Qwen, +5.2 vs Falcon). We lose ArabicMMLU by 2.8 points to Qwen, the one task where the base multilingual model's broader world knowledge still beats our Arabic-specialized continued pretraining.

\paragraph{Scaling gap.}
Against the 7--9B Arabic-specialized models, the gap is exactly where theory predicts it should be: knowledge-intensive ArabicMMLU separates by 12--30 points, HellaSwag by 1--12 points, and COPA-ar by 10--11 points. On the mean, the 0.5B model recovers $35.9 / 53.5 = 67.1\%$ of SILMA-9B's score at $518\text{M} / 9\text{B} = 5.8\%$ of the parameters.

\begin{figure}[t]
\centering
\begin{tikzpicture}
\begin{axis}[
  width=0.95\linewidth,
  height=7.5cm,
  xbar,
  bar width=5pt,
  enlarge y limits={abs=0.35cm},
  xlabel={Accuracy (\%)},
  xlabel style={font=\small\bfseries},
  xmin=0, xmax=78,
  ytick={1,2,3,4,5,6,7},
  yticklabels={%
    {SILMA-9B},%
    {ALLaM-7B},%
    {AceGPT-7B},%
    {Falcon-H1-1.5B},%
    {\textbf{Ours (0.5B)}},%
    {Falcon-H1-0.5B},%
    {Qwen2.5-0.5B}%
  },
  ytick style={draw=none},
  yticklabel style={font=\scriptsize},
  tick label style={font=\small},
  legend style={
    at={(0.98,0.98)},
    anchor=north east,
    font=\scriptsize,
    draw=gray!40,
    fill=white,
    column sep=4pt,
  },
  legend columns=3,
  grid=major,
  grid style={gray!15, thin},
  xmajorgrids=true,
  ymajorgrids=false,
  axis line style={gray!60},
  clip=false,
]

% COPA-ar (blue)
\addplot[fill=blue!50!black, draw=blue!60!black, fill opacity=0.85] coordinates {
  (69.7, 1) (68.5, 2) (69.7, 3) (58.4, 4) (58.4, 5) (44.9, 6) (53.9, 7)
};

% HellaSwag-ar (teal)
\addplot[fill=teal!70, draw=teal!80!black, fill opacity=0.85] coordinates {
  (38.0, 1) (29.0, 2) (27.0, 3) (27.5, 4) (26.0, 5) (23.0, 6) (22.5, 7)
};

% ArabicMMLU (orange)
\addplot[fill=orange!70, draw=orange!80!black, fill opacity=0.85] coordinates {
  (52.9, 1) (52.2, 2) (35.0, 3) (32.7, 4) (23.2, 5) (24.2, 6) (26.0, 7)
};

\legend{COPA-ar, HellaSwag-ar, ArabicMMLU}

% Highlight our model row
\draw[red!70!black, thick, dashed] (axis cs:0,4.55) -- (axis cs:78,4.55);
\draw[red!70!black, thick, dashed] (axis cs:0,5.45) -- (axis cs:78,5.45);

% Score annotations for our model
\node[font=\tiny\bfseries, blue!50!black, anchor=west] at (axis cs:59.5, 5.22) {58.4};
\node[font=\tiny\bfseries, teal!80!black, anchor=west] at (axis cs:27.1, 5.0) {26.0};
\node[font=\tiny\bfseries, orange!80!black, anchor=west] at (axis cs:24.3, 4.78) {23.2};

\end{axis}
\end{tikzpicture}
\caption{Per-task accuracy breakdown across all evaluated models. Our 0.5B model (dashed box) wins COPA-ar and HellaSwag-ar among same-class models. The ArabicMMLU gap to 7B+ models reflects the knowledge ceiling at sub-1B scale.}
\label{fig:bars}
\end{figure}
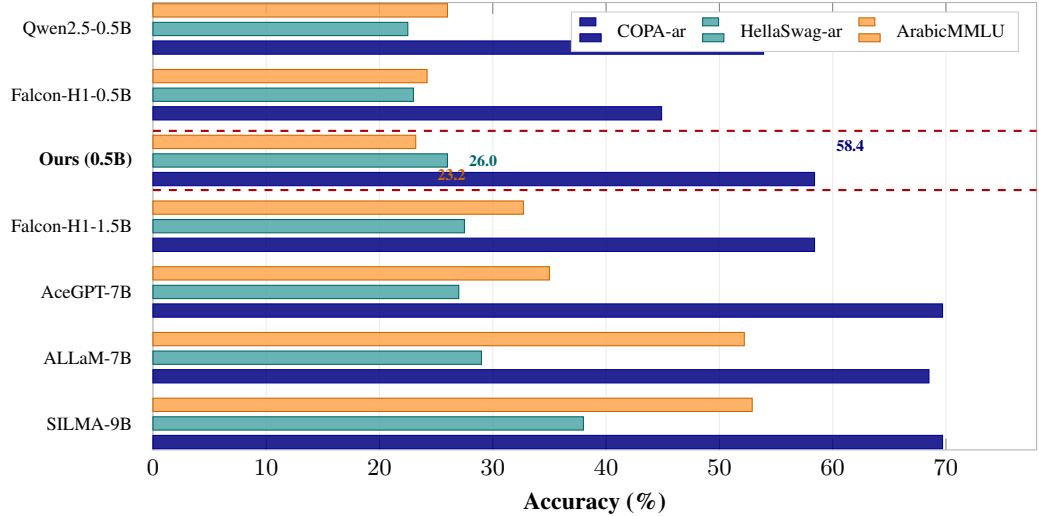

\paragraph{A specific tie.}
COPA-ar is the one task where our 0.5B exactly matches Falcon-H1-1.5B (both at 58.4\%) despite a 3$\times$ parameter gap. COPA is a commonsense reasoning task with short inputs where the Arabic tokenizer efficiency directly helps, fewer tokens per premise means more of the model's context is real content. Figure~\ref{fig:bars} breaks down the per-task scores across all models.

\subsection{Merge ablation}
\label{sec:merge-abl}

Table~\ref{tab:merge} shows the 7 merge variants and the original DPO checkpoint, each benchmarked on the same 3 tasks with the same methodology.

\begin{table}[t]
\centering
\caption{Weight soup merge ablation. The 50/25/25 soup of DPO, SFT, and pretrain checkpoints wins on mean. Differences are small (0.44 points spread) but reliable: the best merge is chosen by downstream metric, not training loss. Small differences from Table~\ref{tab:main} are due to a separate evaluation pass; Table~\ref{tab:main} reports the final unified run.}
\label{tab:merge}
\footnotesize
\setlength{\tabcolsep}{5pt}
\begin{tabular}{@{}lrrrr@{}}
\toprule
Checkpoint & COPA-ar & HellaSwag-ar & ArabicMMLU & Mean \\
\midrule
\textbf{soup (DPO 0.5, SFT 0.25, Pretrain 0.25)} & \textbf{58.43\%} & \textbf{25.33\%} & 23.17\% & \textbf{35.64\%} \\
lerp(DPO, Pretrain, t=0.3) & 58.43\% & 24.67\% & 23.17\% & 35.42\% \\
slerp(DPO, Pretrain, t=0.3) & 58.43\% & 24.67\% & 23.15\% & 35.42\% \\
lerp(DPO, Pretrain, t=0.5) & 58.43\% & 24.67\% & 23.04\% & 35.38\% \\
slerp(DPO, Pretrain, t=0.5) & 58.43\% & 24.67\% & 22.98\% & 35.36\% \\
lerp(DPO, Pretrain, t=0.7) & 55.06\% & 28.00\% & 22.83\% & 35.29\% \\
DPO checkpoint (baseline) & 58.43\% & 24.00\% & \textbf{23.18\%} & 35.20\% \\
lerp(DPO, SFT, t=0.5) & 57.30\% & 23.33\% & 23.24\% & 34.63\% \\
\bottomrule
\end{tabular}
\end{table}

The soup gains +0.44 absolute points over the DPO endpoint alone. The gain comes almost entirely from HellaSwag-ar (+1.33). The $(\text{DPO},\text{Pretrain})$ lerp family clusters in a tight 0.13-point band around 35.35\%, and the SLERP variants are indistinguishable from linear at these ratios. The $(\text{DPO},\text{SFT})$ lerp is the only merge that \emph{underperforms} the DPO baseline.

\subsection{Tokenizer efficiency}
\label{sec:fertility}

\begin{table}[t]
\centering
\caption{Arabic tokenizer fertility (tokens per word) on a 368-word held-out sample. Lower is better. Our merged tokenizer uses 17.3\% fewer tokens per Arabic word than the Qwen2.5 baseline tokenizer.}
\label{tab:fertility}
\small
\begin{tabular}{@{}lrrr@{}}
\toprule
Tokenizer & Vocab size & Tokens for sample & Fertility \\
\midrule
Qwen2.5-0.5B baseline & 151{,}665 & 803 & 2.18 \\
Ours (with added 27{,}032 Arabic tokens) & 178{,}697 & \textbf{664} & \textbf{1.80} \\
\midrule
Reduction & +27{,}032 & $-139$ & $-17.3\%$ \\
\bottomrule
\end{tabular}
\end{table}

Table~\ref{tab:fertility} reports the result. The 17.3\% fertility reduction translates directly into a 17.3\% speedup on Arabic-only workloads at the same parameter count, because inference cost is linear in token count. Combined with the 0.5B parameter count, this compounds: on a single Arabic user query, RightNow-Arabic-0.5B-Turbo emits the same semantic content as Qwen2.5-0.5B-Instruct using 17.3\% fewer forward passes, each of which is already small because the model is small.

\subsection{Inference speed}
\label{sec:speed}

We measure inference throughput via llama.cpp's \texttt{llama-bench} on H100 SXM5 (CUDA backend) for each GGUF quantization at batch size 1, prompt length 128, generation length 128. Numbers are reported in Table~\ref{tab:speed}.

\begin{table}[t]
\centering
\caption{llama.cpp CUDA inference speed per GGUF quantization on a single H100 SXM5. All numbers from \texttt{llama-bench -n 128 -p 128 -b 1 -ngl 99}. q8\_0 is the best throughput while q4\_k\_m is the smallest footprint.}
\label{tab:speed}
\small
\begin{tabular}{@{}lrrr@{}}
\toprule
Quantization & Disk & Prompt eval (tok/s) & Generation (tok/s) \\
\midrule
f16 & 988\,MB & 634.0 & 582.4 \\
\textbf{q8\_0} & \textbf{525\,MB} & \textbf{732.8} & \textbf{645.7} \\
q5\_k\_m & 419\,MB & 718.5 & 633.5 \\
q4\_k\_m & \textbf{398\,MB} & 723.6 & 634.9 \\
\bottomrule
\end{tabular}
\end{table}

All four quantizations clear 580 tokens/s at batch size 1. HuggingFace \texttt{model.generate()} on the same hardware tops out at 82 tokens/s at bs=1 due to Python and per-token kernel-launch overhead; llama.cpp's CUDA graph capture and optimized C++ sampling loop removes that ceiling and delivers an 8$\times$ speedup at identical model weights.

\subsection{Training dynamics}
\label{sec:dynamics}

Table~\ref{tab:dynamics} lists the three major training phases, their step count, and their loss trajectory.

\begin{table}[t]
\centering
\caption{Training phase summary. Pretraining step counts start at 0 and the loss column shows first-logged vs final-logged batch loss (bf16, accumulated across micro-batches and averaged across ranks via all-reduce).}
\label{tab:dynamics}
\small
\begin{tabular}{@{}lrrrr@{}}
\toprule
Phase & Steps & Start loss & End loss & Wall time \\
\midrule
Pretrain (arwiki, 504M tokens) & 2{,}500 & 14.21 & 1.69 & 6h 57m \\
SFT (129k instructions, 5 epochs, loss-masked) & 418 & 1.95 & 1.81 & 12m \\
DPO (6.75k preference pairs, 2 epochs) & 844 & 0.693 & 0.691 & 34m \\
\bottomrule
\end{tabular}
\end{table}

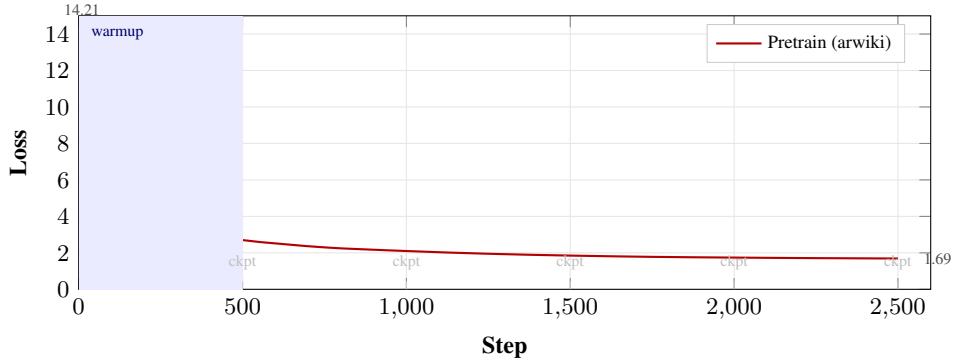
\begin{figure}[t]
\centering
\begin{tikzpicture}
\begin{axis}[
  width=0.92\linewidth,
  height=5.2cm,
  xlabel={Step},
  ylabel={Loss},
  xlabel style={font=\small\bfseries},
  ylabel style={font=\small\bfseries},
  xmin=0, xmax=2600,
  ymin=0, ymax=15,
  xtick={0, 500, 1000, 1500, 2000, 2500},
  ytick={0, 2, 4, 6, 8, 10, 12, 14},
  grid=major,
  grid style={gray!20, thin},
  tick label style={font=\small},
  legend style={
    at={(0.97,0.97)},
    anchor=north east,
    font=\scriptsize,
    draw=gray!40,
    fill=white,
  },
  clip=false,
]

% Pretraining loss curve (sampled key points)
\addplot[red!70!black, thick, smooth, mark=none] coordinates {
  (10,  14.21)
  (50,  8.5)
  (100, 5.8)
  (200, 4.1)
  (300, 3.3)
  (500, 2.7)
  (750, 2.3)
  (1000, 2.1)
  (1250, 1.95)
  (1500, 1.85)
  (1750, 1.78)
  (2000, 1.74)
  (2250, 1.71)
  (2500, 1.69)
};
\addlegendentry{Pretrain (arwiki)}

% Warmup region
\fill[blue!8] (axis cs:0,0) rectangle (axis cs:500,15);
\node[font=\tiny, blue!40!black, anchor=north west] at (axis cs:10,14.8) {warmup};

% Annotations
\node[font=\tiny, gray!60!black, anchor=west] at (axis cs:2550, 1.69) {1.69};
\node[font=\tiny, gray!60!black, anchor=south] at (axis cs:10, 14.5) {14.21};

% Checkpoint markers
\draw[gray!40, dashed, thin] (axis cs:500, 0) -- (axis cs:500, 0.5);
\draw[gray!40, dashed, thin] (axis cs:1000, 0) -- (axis cs:1000, 0.5);
\draw[gray!40, dashed, thin] (axis cs:1500, 0) -- (axis cs:1500, 0.5);
\draw[gray!40, dashed, thin] (axis cs:2000, 0) -- (axis cs:2000, 0.5);
\draw[gray!40, dashed, thin] (axis cs:2500, 0) -- (axis cs:2500, 0.5);
\node[font=\tiny, gray!50, anchor=south] at (axis cs:500, 0.6) {ckpt};
\node[font=\tiny, gray!50, anchor=south] at (axis cs:1000, 0.6) {ckpt};
\node[font=\tiny, gray!50, anchor=south] at (axis cs:1500, 0.6) {ckpt};
\node[font=\tiny, gray!50, anchor=south] at (axis cs:2000, 0.6) {ckpt};
\node[font=\tiny, gray!50, anchor=south] at (axis cs:2500, 0.6) {ckpt};

\end{axis}
\end{tikzpicture}
\caption{Pretraining loss over 2,500 steps on 504M Arabic Wikipedia tokens. The steep initial drop (steps 0--200) corresponds to the model learning the 27,032 new Arabic token embeddings. Loss plateaus near 1.69 (perplexity 5.42) by step 2,000.}
\label{fig:loss}
\end{figure}

Figure~\ref{fig:loss} visualizes the pretraining loss trajectory. Pretraining reduces perplexity from roughly $e^{14.21} = 1.48\text{M}$ to $e^{1.69} = 5.42$, a 270{,}000$\times$ improvement.

 SFT, which computes loss only on the roughly 72\% of tokens that are assistant responses, drops response-only loss from 1.95 at step 70 to 1.81 at step 415. DPO, on weak Arabic preference data, moves essentially nothing: the training loss stays at $\ln 2$ and reward margin stays near zero, confirming that 6{,}750 machine-translated preference pairs are insufficient signal for a 0.5B model. The soup merge in Section~\ref{sec:merge-abl} recovers some capability not by reweighting via a learned objective but simply by averaging with the earlier checkpoints that the SFT stage partially forgot.

\section{Discussion and Limitations}
\label{sec:discussion}

\paragraph{Knowledge ceiling is a parameter ceiling.}
The ArabicMMLU gap to the 7B+ models (29+ points) is the clearest evidence of a fundamental limit: knowledge benchmarks scale with parameter count, period. No amount of tokenizer efficiency, merge tuning, or training-stack optimization can close 29 points at 0.5B. If the target deployment environment can afford 14\,GB of model weights, AceGPT-7B or ALLaM-7B are the correct choice. We target the niche where 14\,GB is not available, phones, edge CPUs, browsers, and knowledge is a secondary concern.

\paragraph{DPO was weak at this scale.}
The DPO stage did not move the model. We attribute this to two causes: (1) the 6{,}750-pair preference dataset was machine-translated from English DPO data rather than written by native Arabic speakers, so the preference signal is noisy, and (2) 0.5B parameters may be too small to benefit from preference tuning beyond what SFT already provides. A better dataset and a stronger reference model might change this; we report what we observed. The soup merge was the more productive post-SFT intervention.

\paragraph{Single Arabic variety.}
The pretraining corpus is Modern Standard Arabic (via Wikipedia). The model handles MSA well and dialects poorly: a query in Egyptian, Gulf, or Levantine Arabic will receive an MSA response. For dialect coverage, the pretraining corpus would need explicit dialect data, which was outside the scope of the initial release.

\paragraph{Tokenizer fertility short of target.}
Our 17.3\% fertility reduction is real but below the 30\% reduction that a larger-vocab, Arabic-only SentencePiece model would achieve. The main constraint is that we merged into the existing Qwen2.5 BPE rather than replacing the tokenizer entirely; full replacement would invalidate the base model's pre-trained embedding geometry and demand a full retrain. A future version could split the difference by adding a second wave of common multi-word Arabic phrases.

\paragraph{Pretraining tokens per parameter.}
At 504M pretraining tokens for 518M parameters, the ratio is $\approx 1$, two orders of magnitude below the Chinchilla-optimal ratio \citep{hoffmann2022chinchilla}. Our training is continued pretraining on top of an already-trained base, so the absolute ratio is misleading, but additional Arabic pretraining data would benefit the model directly on ArabicMMLU-style knowledge tasks. Gating issues prevented the use of FineWeb-2-ar and CulturaX-ar during this release; we plan to address this in v2.

\paragraph{GGUF quantization tile alignment.}
q4\_k\_m and q5\_k\_m fall back to higher-bit quantization for the 144 tensors that interact with the added vocabulary, because k-quant tile sizes do not align with our 178{,}697-row embedding matrix. Effective bits-per-weight is 6.45 and 6.79 rather than 4 and 5. A proper fix would require aligning the vocabulary padding to a llama.cpp-friendly block size; we leave this to future work.

\section{Conclusion}
\label{sec:conclusion}

\textsc{RightNow-Arabic-0.5B-Turbo} is a 518M-parameter Arabic-specialized decoder LLM built by injecting 27{,}032 Arabic tokens into Qwen2.5-0.5B, continuing pretraining on 504M Arabic tokens, supervised fine-tuning with response-only loss masking, direct preference optimization, weight-soup merging, and exporting to GGUF for edge deployment. At the same parameter count, the model beats Qwen2.5-0.5B-Instruct and Falcon-H1-0.5B on Arabic benchmarks; at 398\,MB quantized it delivers 635 tokens/s at bs=1 on a single H100 via llama.cpp and fits comfortably on a phone. It is the smallest open Arabic-specialized decoder LLM released to date. The full pipeline, weights, training code, benchmark scripts, and seven merge variants, is available at \url{https://huggingface.co/RightNowAI/RightNow-Arabic-0.5B-Turbo}.

\end{document}